\documentclass[sigconf]{acmart}
\renewcommand\footnotetextcopyrightpermission[1]{} 
\AtBeginDocument{%
  \providecommand\BibTeX{{%
    \normalfont B\kern-0.5em{\scshape i\kern-0.25em b}\kern-0.8em\TeX}}}

\acmConference[Sciknow '19]{Sciknow '19: Third International Workshop on Capturing Scientific Knowledge}{November 19, 2019}{Los Angeles, California, USA}

\usepackage{array,multirow,graphicx}
\begin{document}
\thanks{Copyright \textcopyright 2019 for this paper by its authors. Use permitted under Creative Commons License Attribution 4.0 International (CC BY 4.0).}

\title{Semantic Workflows and Machine Learning for the Assessment of Carbon Storage by Urban Trees}


\author{Juan Carrillo}
\affiliation{%
  \institution{University of Waterloo}
  \city{Waterloo}
  \country{Canada}
}
\email{jmcarril@uwaterloo.ca}

\author{Daniel Garijo}
\affiliation{%
  \institution{Information Sciences Institute}
  \city{University of Southern California}
  \country{USA}
}
\email{dgarijo@isi.edu}

\author{Mark Crowley}
\affiliation{%
  \institution{University of Waterloo}
  \city{Waterloo}
  \country{Canada}
}
\email{mark.crowley@uwaterloo.ca}

\author{Rober Carrillo}
\affiliation{%
  \institution{United World Colleges}
  \city{Bogota}
  \country{Colombia}
}
\email{rober.carrillo@uwcim.net}

\author{Yolanda Gil}
\affiliation{%
  \institution{Information Sciences Institute}
  \city{University of Southern California}
  \country{USA}
}
\email{gil@isi.edu}

\author{Katherine Borda}
\affiliation{%
  \city{Waterloo}
  \country{Canada}
}
\email{kbordac@gmail.com}

\renewcommand{\shortauthors}{Carrillo, et al.}

\begin{abstract}

Climate science is critical for understanding both the causes and consequences of changes in global temperatures and has become imperative for decisive policy-making. However, climate science studies commonly require addressing complex interoperability issues between data, software, and experimental approaches from multiple fields. Scientific workflow systems provide unparalleled advantages to address these issues, including reproducibility of experiments,  provenance capture, software reusability and knowledge sharing. In this paper, we introduce a novel workflow with a series of connected components to perform spatial data preparation, classification of satellite imagery with machine learning algorithms, and assessment of carbon stored by urban trees. To the best of our knowledge, this is the first study that estimates carbon storage for a region in Africa following the guidelines from the Intergovernmental Panel on Climate Change (IPCC).

\end{abstract}




\maketitle

\section{Introduction}

Climate science requires modeling natural and man-made processes that are highly complex, exhibit non-linear dynamics and possess disparate spatial and temporal scales. Handling this complexity requires a holistic approach among multiple disciplines \cite{meadow2015moving}, but scientists from different fields may also need to use domain-specific data sources, methods, and computational models. The integration of their knowledge and experiments is a challenging task \cite{Daron2015}, especially when a study is expected to provide actionable insights for decision making at regional and local scale.


Scientific workflows have emerged as an integrated solution to manage this challenge, as they capture the computational steps and data dependencies required to carry out a computational experiment \cite{Taylor:2014}. Scientific workflows ease  data handling (metadata, provenance), component versioning (parametrization, calibration) and have a clear separation between workflow design and workflow execution \cite{BARSEGHIAN201042} \cite{wolstencroft2013taverna}. One of the major advantages of scientific workflow systems is their role in improving the reproducibility of scientific studies. Reproducibility plays a critical role in climate sciences due to their impact in our society  \cite{wired}. In fact, due to issues with the documentation of experiments, some climate science studies have been re-examined lately due to their impact in global policy-making \cite{gillingham2018modeling} and  water resources management \cite{kundzewicz2018uncertainty}.   



In this paper we describe the process we followed to design and implement reusable scientific workflows for the climate sciences. In particular, we focus on carbon storage assessment by using urban trees, a common requirement for cities to reduce their carbon emissions globally. Our contributions include the development of a library of components for preparing geospatial data by doing coordinate transformations, the integration of machine learning components to classify trees in satellite images and the creation of workflows for carbon storage assessment in cities. In order to implement these workflows, we use the Workflow Instance Generation and Selection (WINGS) system \cite{gil-etal-ieee-is-11}, which has been successfully used for applications in domains ranging from Genomics \cite{gil-etal-ismb12} to Geosciences \cite{gil-etal-agu12}.

The paper starts by giving an overview of previous research on the assessment of carbon storage by urban trees according to the guidelines published by the Intergovernmental Panel on Climate Change (IPCC), highlighting the advantages and limitations of the most recent methods. We then describe the design considerations of our scientific workflows as well as their experimental implementation and evaluation. The paper continues with a discussion of results and suggested future work.

\section{Background: Carbon emissions and storage}

There is an increasing interest among scientific organizations and national governments regarding Carbon emissions and their role in climate science \cite{figueres2017three}. The consequences of higher concentrations of carbon gases in the atmosphere are now more explicit and international organizations such as the Intergovernmental Panel on Climate Change (IPCC) are leading initiatives to monitor national efforts to lower emissions and increase carbon storage \cite{edenhofer2015climate}. Monitoring carbon emissions is fundamental to inform government policies in topics such as renewable energy, transportation, and manufacturing technologies. Similarly, the assessment of carbon storage is equally important to guide hazard mitigation efforts \cite{gough2016carbon}.

Most studies in climate science require domain knowledge to design and run the experiments \cite{incropera2016climate}. But the increasing concerns about the changing climate require strategies to streamline the replication of assessment studies, the reusability of data, methods, and results. 
The use of scientific workflows can significantly improve the implementation and reproducibility of carbon assessment studies, with additional gains in data and model sharing as well as knowledge capture through semantic representations.

\subsection{Carbon assessment by urban trees}

Urban trees provide a natural and cost-effective alternative to capture and store carbon in cities. Having trees in densely populated areas also improve human health and biodiversity and provide benefits for flood prevention and reduced cooling costs, among other benefits \cite{livesley2016urban}. In 2003,  the IPCC published the \textit{Good Practice Guidance for Land Use, Land-Use Change and Forestry} \cite{penman2003good} and in 2006 the \textit{IPCC Guidelines for National Greenhouse Gas Inventories} \cite{eggleston2006}. These guidelines suggest the use of area covered by trees, shrubs, and herbaceous (perennial) plants to determine the amount of carbon stored as biomass in settlements. However, due to the limited availability of detailed data the majority of published studies focus only on tree cover. While these two documents describe the stages of an assessment study, including aspects such as data collection and uncertainty estimation, they suggest governments to deal with minor implementation details according to their technical capacity and available resources.

Published carbon assessment studies have used different combinations of data, methods, and software. These studies can be divided into two major groups according to the data collection approaches and models they use. Assessments in the first group use a statistics point sampling technique to estimate tree density from aerial imagery \cite{mcgovern2016canadian}\cite{pasher2014assessing}\cite{nowak2013carbon}. This method is easy to implement and only requires imagery for sample areas, but the outcome is a percentage value that does not describe the spatial distribution of trees. The second group of methods use LiDAR, aerial or satellite imagery to provide a comprehensive assessment of tree coverage, including their spatial distribution \cite{tigges2017modeling}\cite{schreyer2014using}\cite{davies2011mapping}\cite{raciti2014mapping}. However, the second approach requires imagery for the complete the area of study as well as the configuration of more complex methods for detection or classification of urban trees. Later in this document we introduce our own method, which belongs to the second group and uses freely available satellite imagery and a carefully designed workflow to facilitate implementation and reuse.

One common limitation of carbon assessment studies is the lack of a systematic approach to share data, models, software, and results. Regardless of the specific data source, technology, or processing method, most reports only contain descriptions of the work done, which are not enough to replicate the experiments or reuse the software tools. It is difficult for other scientists, especially in developing countries, to translate the information from those reports into actionable knowledge that serves them to efficiently incorporate published research into their own studies.

\section{Scientific Workflow Design}

With the advantages and limitations of published methods in mind, we design a new workflow to efficiently determine tree cover for urban areas 
We start by presenting the advantages of knowledge capture systems to represent models in geosciences as scientific workflows and then describe how we leverage previous research on carbon assessment and tree mapping to design our own workflow.





\begin{figure}[h]
  \centering
  \includegraphics[width=\linewidth]{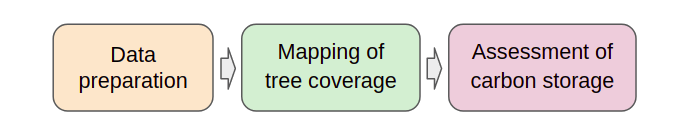}
  \caption{Stages of our carbon assessment workflow}
  \Description{Stages of our carbon assessment workflow}
  \label{fig:c-assess-wf}
\end{figure}

The workflow is designed as multiple interconnected components in WINGS that operate in three consecutive stages as seen in Figure ~\ref{fig:c-assess-wf}, data preprocessing, mapping of tree coverage, and assessment of carbon storage. Our workflow is based on previous work in which high resolution satellite imagery are used to produce land cover maps over urban areas \cite{raciti2014mapping} \cite{schreyer2014using}. However, we focus on using freely available medium resolution satellite imagery from the Sentinel-2 sensors \cite{drusch2012sentinel} to facilitate replication by other researchers.

Spatial data preprocessing stage involves common operations for experiments across earth sciences, such as transformation of coordinate systems and conversion between file formats. We have designed all these data preprocessing  steps as reusable building blocks so they can be included in other workflows. These preprocessing operations are also known as Extract Transform Load (ETL) tasks and are implemented using the Geospatial Data Abstraction Library GDAL \cite{gdalcite}.

We map the tree coverage using satellite image classification and design multiple components to train machine learning algorithms, classify the image over an area of interest, produce a visualization ready tree cover map, and determine the resulting  accuracy. The Machine Learning algorithms we implement and calibrate are Random Forests and Support Vector Machines, both available in the Orfeo Remote Sensing toolbox \cite{grizonnet2017orfeo}. We train the algorithms using sample points collected through visual inspection. 

\begin{figure}[h]
  \centering
  \includegraphics[width=\linewidth]{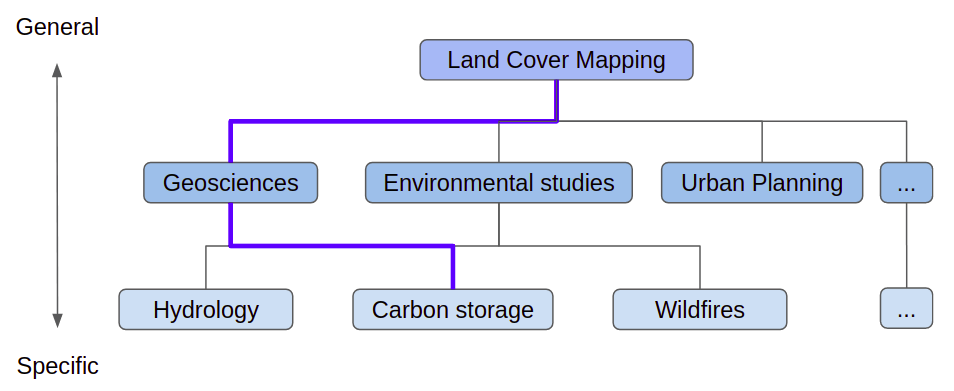}
  \caption{Some applications of land cover mapping}
  \Description{Some applications of land cover mapping}
  \label{fig:landcover-mapping}
\end{figure}

Our workflow generates a map that includes other land cover categories such as water, grass and built areas; which may additionally serve for studies in hydrology, planning, and other applications (as seen in Figure ~\ref{fig:landcover-mapping}). Moreover, this map is generated in a standard format for further use in Geographic Information Systems or other scientific platforms.

The assessment of carbon storage is completed following the IPCC guidelines to calculate carbon stored based on urban canopy area. In the calculation we multiply the canopy area by a conversion factor to estimate carbon stored in the form of biomass. Since no values are published specifically for Africa (our initial region of interest) we use a default value suggested by the IPCC.

\begin{figure}[h]
  \centering
  \includegraphics[width=\linewidth]{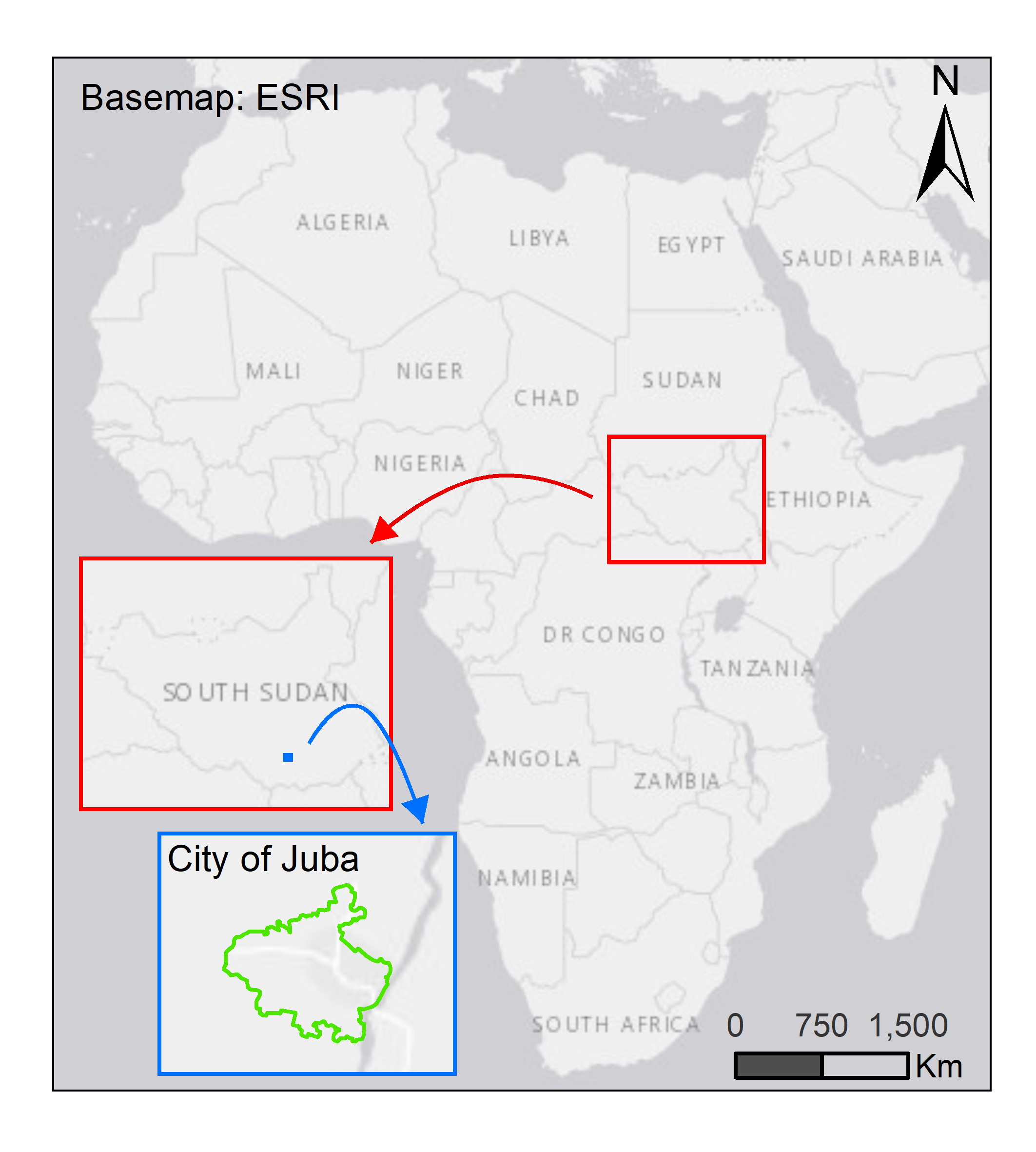}
  \caption{City of Juba in South Sudan, Africa}
  \Description{Juba_in_Africa}
  \label{fig:Juba_in_Africa}
\end{figure}

\section{Area of Study and Data}

Our area of study is the city of Juba in South Sudan. It is the current capital of the country and serves as its main commercial and transportation hub with an estimated population of nearly 386,000 inhabitants \cite{ciagency_2018}. Juba is located in the southern region of the country and has an extension of 103 km\textsuperscript{2} according to the urban boundary retrieved in July 2019 from Open Street Maps \cite{haklay2008openstreetmap}. Figure ~\ref{fig:Juba_in_Africa} shows the location of Juba and South Sudan in the African continent.

The country of South Sudan currently faces multiple issues, including political instability \cite{sorbo2013sudan}, poor health services \cite{macharia2017spatial}, and a lack of infrastructure, especially for storage and distribution of water \cite{nassif2016infrastructure}. Few mapping projects have been conducted in Africa, and especially in South Sudan \cite{baariu2019state}\cite{kagawa2018mapping}, which hinder the work of humanitarian and non government organizations; therefore, any study aiming to understand the characteristics of the rural or urban territories will likely produce positive outcomes in the short and long term.

The data we use includes 4000 sample points digitized through visual inspection from Google Earth high resolution satellite imagery. The sample points are spatially dispersed across the city and distributed in four categories, tree canopy, grass, impervious, and water, with 1000 points per category to have a balanced dataset. Impervious includes all built or bare areas not covered by vegetation, water, or agriculture. Examples of impervious areas are constructions, parking lots, roads, airport runways, and rocky surfaces.

Additionally, we use a multi-spectral satellite image from the Sentinel-2 sensor made available by the European Space Agency (ESA) \cite{drusch2012sentinel}. While this image originally includes 13 bands we use only the four bands with a pixel size equivalent to 10 meters on the ground. The first three bands collect data in the visible light spectrum and the fourth in the near infrared spectrum, with the latter being particularly useful for detecting vegetation.

\begin{figure}[h]
  \centering
  \includegraphics[width=\linewidth]{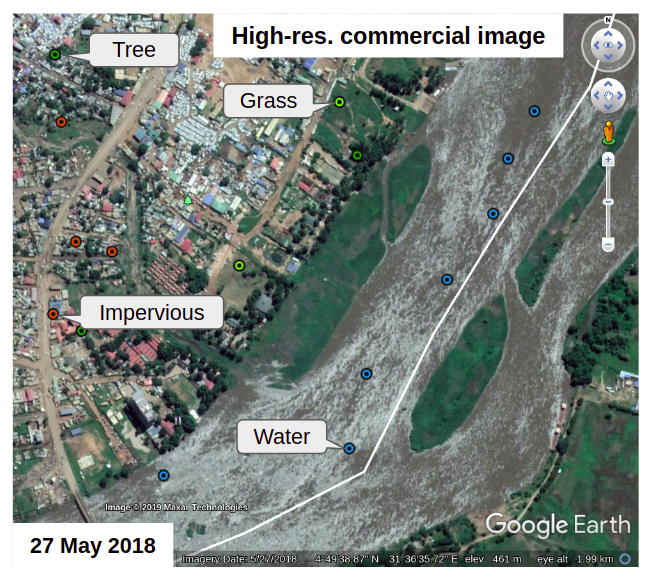}
  \includegraphics[width=\linewidth]{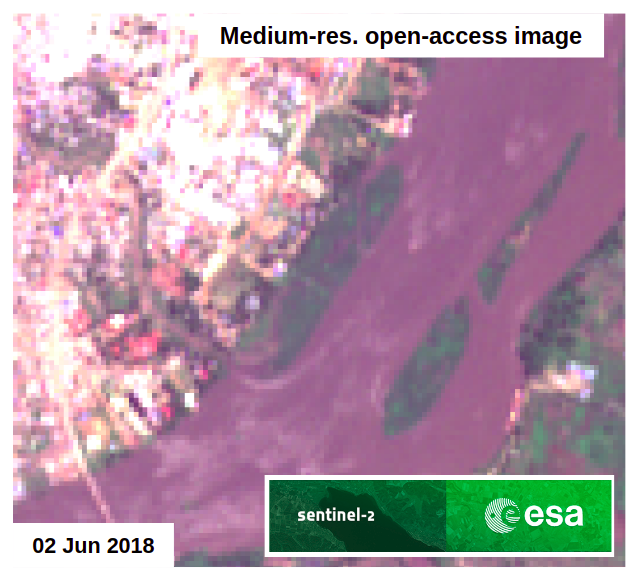}
  \caption{Sample points and satellite imagery of Juba}
  \Description{Sample points and satellite imagery of Juba}
  \label{fig:high-and-med-res}
\end{figure}

Figure ~\ref{fig:high-and-med-res} shows some sample points of the four land cover categories in a neighborhood close to the White Nile river in the city of Juba, overlaid on high resolution satellite imagery from Google Earth and the same area as seen on the medium resolution Sentinel-2 image.

\section{Implementation and Results}

The workflow is implemented as a series of software components in WINGS.  Additionally, each component is designed to run a particular task and configured with relevant parameters to calibrate their functionality. Components are coded as Python scripts that serve as a high level interface for us to use libraries such as GDAL for data preprocessing and Orfeo command line tools for satellite image classification.

These components are connected through intermediate datasets that in turn are outputs and inputs for the previous and following components. We created 14 components in total, with eight dedicated for data preparation, five for mapping of tree coverage, and one for carbon assessment. The use of components as modular pieces of software to accomplish specific data processing tasks creates opportunities for reusability across a variety of models in Earth Sciences and Geospatial technologies.

\begin{figure}[h]
  \centering
  \includegraphics[width=0.75\linewidth]{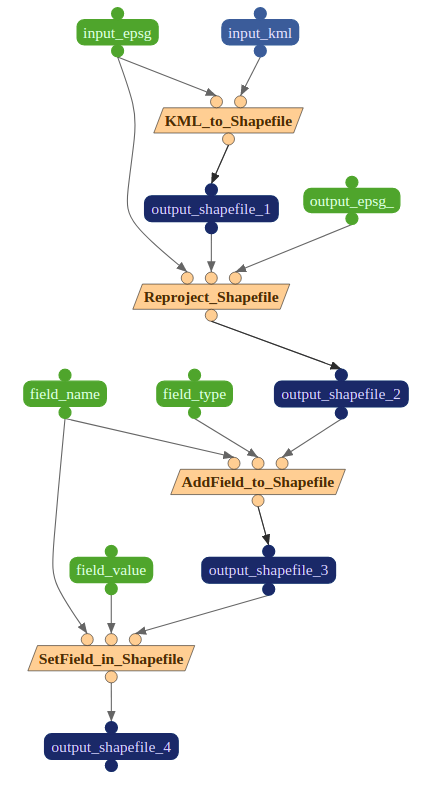}
  \caption{Workflow fragment for data preparation of sample points}
  \Description{Workflow fragment for data preparation of sample points}
  \label{fig:sample-points-prep}
\end{figure}

The eight components designed for data preparation allow researchers to handle datasets in the most common file formats and transform them according to the particular goals of the study. Some operations correspond to sub-setting the spatial extent of the data, changing its coordinate reference system, and editing the attributes for tabular datasets. When running the workflow for a new area of study, the task of setting parameters such as the coordinate reference system is facilitated by the WINGS system, which suggest the most appropriate value according to the geographic region as configured by the workflow designer. Figure ~\ref{fig:sample-points-prep} shows a fragment of our workflow where we use multiple data preparation components to perform a format transformation and reprojecting a file.

\begin{figure}[h]
  \centering
  \includegraphics[width=\linewidth]{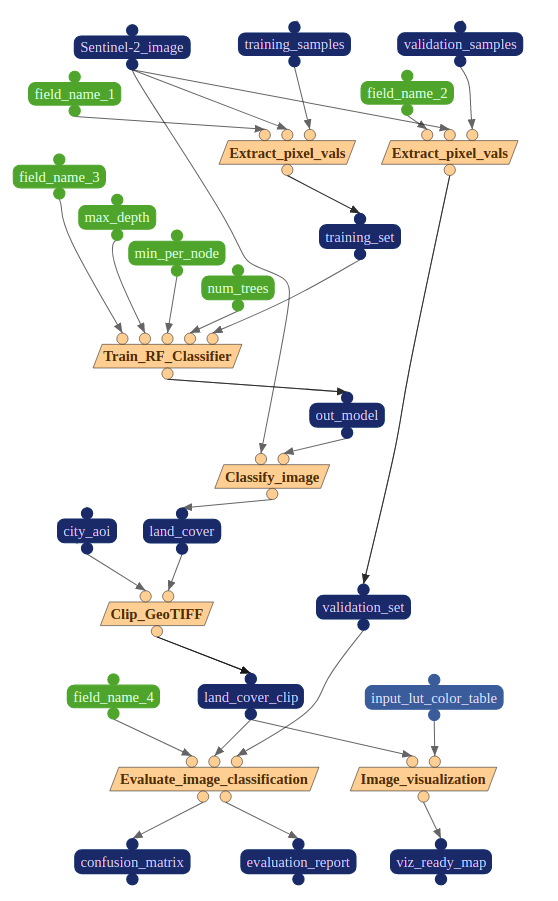}
  \caption{Workflow fragment for mapping tree cover}
  \Description{Workflow fragment for mapping tree cover}
  \label{fig:mapping-tree-cover}
\end{figure}

Mapping of tree coverage includes four components focused on training the Machine Learning algorithms to classify the Sentinel-2 satellite image using the sample points digitized through visual inspection. Initially, one component extracts the pixel values of the satellite image for the 4000 sample locations. Next, another component uses 80\% of these values as training and 20\% as validation data to train the Random Forest and Support Vector Machine algorithms. Then we use the trained algorithms to perform pixel level classification of the Sentinel-2 image for the entire area of interest and output the tree cover map. Subsequently, one more component evaluates the classification accuracy. Figure ~\ref{fig:mapping-tree-cover} shows a fragment of our workflow where we use multiple components to map tree cover.

Tables ~\ref{tab:ncf} and ~\ref{tab:cf_wn} show the confusion matrix with and without normalization resulting from the evaluation of the Random Forest classification algorithm. We see that the tree cover category is the one with the lowest accuracy with only 108 out of the 200 trees in the test set correctly classified. This is likely the result of using medium resolution satellite imagery (Sentinel-2) with a ground pixel size of 10 m, in other words,  the canopy area of a tree should be about 100 m\textsuperscript{2} to be easily identifiable in at least one pixel, without considering boundary issues between adjacent pixels. The grass and impervious land cover categories exhibit a comparable accuracy of 65\% and 73\% consequently. While areas corresponding to these two categories show a slightly better accuracy they are still hard to differentiate, presumably due to grass patches and house rooftops with a size smaller than the area of a pixel (100 m\textsuperscript{2}). For the water land cover category the algorithm reaches an almost perfect accuracy, which is anticipated due to the significant difference in the way it reflects the light compared to the other three land cover categories, especially in the near infrared band (B4).

\begin{table}[]
  \caption{Normalized confusion matrix}
  \label{tab:ncf}
\begin{tabular}{cccccc}
\cline{3-6} 
\multirow{4}{*}{\rotatebox{90}{True label}} & \multicolumn{1}{c|}{Trees}      & 0.54   & 0.18    & 0.28         & \multicolumn{1}{c|}{0}   \\
                            & \multicolumn{1}{c|}{Grass}      & 0.32    & 0.65   & 0.14         & \multicolumn{1}{c|}{0}   \\
                            & \multicolumn{1}{c|}{Impervious} & 0.17    & 0.09    & 0.73        & \multicolumn{1}{c|}{0}   \\
                            & \multicolumn{1}{c|}{Water}      & 0.01     & 0     & 0          & \multicolumn{1}{c|}{0.99} \\ \cline{3-6} 
\multicolumn{2}{c}{\multirow{2}{*}{}}                         & Trees\hspace{1mm} & Grass\hspace{1mm} & Impervious\hspace{0mm} & Water\hspace{1mm}                    \\
\multicolumn{2}{c}{}                                          & \multicolumn{4}{c}{Predicted label}                  
\end{tabular}
\end{table}

\begin{table}[]
  \caption{Confusion matrix, without normalization}
  \label{tab:cf_wn}
\begin{tabular}{cccccc}
\cline{3-6} 
\multirow{4}{*}{\rotatebox{90}{True label}} & \multicolumn{1}{c|}{Trees}      & 108   & 37    & 55         & \multicolumn{1}{c|}{0}   \\
                            & \multicolumn{1}{c|}{Grass}      & 42    & 131   & 27         & \multicolumn{1}{c|}{0}   \\
                            & \multicolumn{1}{c|}{Impervious} & 35    & 17    & 147        & \multicolumn{1}{c|}{1}   \\
                            & \multicolumn{1}{c|}{Water}      & 1     & 0     & 0          & \multicolumn{1}{c|}{199} \\ \cline{3-6} 
\multicolumn{2}{c}{\multirow{2}{*}{}}                         & Trees\hspace{1mm} & Grass\hspace{1mm} & Impervious\hspace{0mm} & Water\hspace{1mm}                    \\
\multicolumn{2}{c}{}                                          & \multicolumn{4}{c}{Predicted label}                  
\end{tabular}
\end{table}

\begin{figure}[t!]
  \centering
  \includegraphics[width=\linewidth]{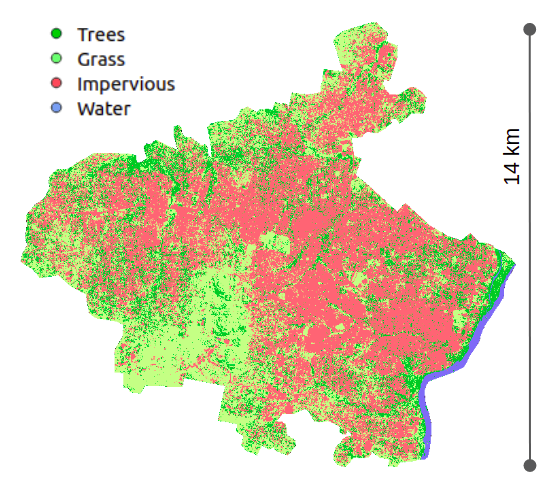}
  \caption{Resulting land cover map for the city of Juba}
  \Description{Resulting land cover map for the city of Juba}
  \label{fig:land-cover-map.png}
\end{figure}

The assessment of carbon storage is performed by one component that receives the tree cover map shown in Figure ~\ref{fig:land-cover-map.png}, calculates the total canopy area and uses a carbon removal factor to determine the total amount of carbon stored as biomass in the city of Juba. We use a default carbon removal factor of 2.9 tonnes C (ha crown cover)\textsuperscript{-1} yr \textsuperscript{-1}, which is the value suggested by the IPCC for Tier 2a studies \cite{eggleston2006}. 
Multiplying the default carbon removal factor by the total tree cover area (10,519 ha) from the previous stage, we calculate that trees in the city of Juba remove 30,506 tonnes C yr \textsuperscript{-1}. This amount is equivalent to the carbon dioxide emitted by 6632 passenger vehicles per year \cite{epa_passenger_emissions}. This value along with the classification accuracy from the previous stage are valuable information for countries to prepare their carbon assessment reports according to the IPCC guidelines. The code used to create the geospatial transformations \footnote{\url{https://github.com/jmcarrillog/geospatial-etl}} and the carbon assessment workflows \footnote{\url{https://github.com/jmcarrillog/machine-learning-workflow-for-carbon-assessment}} is available online. 

\section{Conclusions and Future Work}
In this paper we introduced our work to create a library of workflow components to perform spatial data transformations, land cover mapping and assessment of carbon storage. By leveraging scientific workflows, we aim to ease the reusability of these components in other workflows and the reproducibility and transparency of carbon assessment studies. Our future work will focus on two main areas: first, we aim to test our workflow using data from other locations around the globe, which requires additional training data points. Second, we will focus on calibration of the parameters for the classifiers to improve our classification accuracy during the land cover mapping stage.

\section{Acknowledgments}

Juan Carrillo was financially supported for this project by the Mitacs Globalink Research Award and the University of Waterloo Machine Learning Lab. Juan Carrillo gives special thanks to Mark Crowley, Daniel Garijo, and Yolanda Gil for their mentoring and valuable contributions during this research project.







\bibliographystyle{ACM-Reference-Format}
\bibliography{references}


\begin{thebibliography}{36}


\ifx \showCODEN    \undefined \def \showCODEN     #1{\unskip}     \fi
\ifx \showDOI      \undefined \def \showDOI       #1{#1}\fi
\ifx \showISBNx    \undefined \def \showISBNx     #1{\unskip}     \fi
\ifx \showISBNxiii \undefined \def \showISBNxiii  #1{\unskip}     \fi
\ifx \showISSN     \undefined \def \showISSN      #1{\unskip}     \fi
\ifx \showLCCN     \undefined \def \showLCCN      #1{\unskip}     \fi
\ifx \shownote     \undefined \def \shownote      #1{#1}          \fi
\ifx \showarticletitle \undefined \def \showarticletitle #1{#1}   \fi
\ifx \showURL      \undefined \def \showURL       {\relax}        \fi
\providecommand\bibfield[2]{#2}
\providecommand\bibinfo[2]{#2}
\providecommand\natexlab[1]{#1}
\providecommand\showeprint[2][]{arXiv:#2}

\bibitem[\protect\citeauthoryear{Agency}{Agency}{2018}]%
        {ciagency_2018}
\bibfield{author}{\bibinfo{person}{Central~Intelligence Agency}.}
  \bibinfo{year}{2018}\natexlab{}.
\newblock \bibinfo{title}{The World Factbook: South Sudan}.
\newblock
\newblock
\urldef\tempurl%
\url{https://www.cia.gov/library/publications/the-world-factbook/geos/od.html}
\showURL{%
\tempurl}


\bibitem[\protect\citeauthoryear{Baariu, Mulaku, Siriba, et~al\mbox{.}}{Baariu
  et~al\mbox{.}}{2019}]%
        {baariu2019state}
\bibfield{author}{\bibinfo{person}{Sabina~N Baariu}, \bibinfo{person}{Galcano~C
  Mulaku}, \bibinfo{person}{David~N Siriba}, {et~al\mbox{.}}}
  \bibinfo{year}{2019}\natexlab{}.
\newblock \showarticletitle{State of Cartographic Services among the East
  African Community Member States}.
\newblock \bibinfo{journal}{\emph{Journal of Geographic Information System}}
  \bibinfo{volume}{11}, \bibinfo{number}{01} (\bibinfo{year}{2019}),
  \bibinfo{pages}{56}.
\newblock


\bibitem[\protect\citeauthoryear{Barseghian, Altintas, Jones, Crawl, Potter,
  Gallagher, Cornillon, Schildhauer, Borer, Seabloom, and Hosseini}{Barseghian
  et~al\mbox{.}}{2010}]%
        {BARSEGHIAN201042}
\bibfield{author}{\bibinfo{person}{Derik Barseghian}, \bibinfo{person}{Ilkay
  Altintas}, \bibinfo{person}{Matthew~B. Jones}, \bibinfo{person}{Daniel
  Crawl}, \bibinfo{person}{Nathan Potter}, \bibinfo{person}{James Gallagher},
  \bibinfo{person}{Peter Cornillon}, \bibinfo{person}{Mark Schildhauer},
  \bibinfo{person}{Elizabeth~T. Borer}, \bibinfo{person}{Eric~W. Seabloom},
  {and} \bibinfo{person}{Parviez~R. Hosseini}.}
  \bibinfo{year}{2010}\natexlab{}.
\newblock \showarticletitle{Workflows and extensions to the Kepler scientific
  workflow system to support environmental sensor data access and analysis}.
\newblock \bibinfo{journal}{\emph{Ecological Informatics}} \bibinfo{volume}{5},
  \bibinfo{number}{1} (\bibinfo{year}{2010}), \bibinfo{pages}{42 -- 50}.
\newblock
\showISSN{1574-9541}
\urldef\tempurl%
\url{https://doi.org/10.1016/j.ecoinf.2009.08.008}
\showDOI{\tempurl}
\newblock
\shownote{Special Issue: Advances in environmental information management.}


\bibitem[\protect\citeauthoryear{Daron, Sutherland, Jack, and Hewitson}{Daron
  et~al\mbox{.}}{2015}]%
        {Daron2015}
\bibfield{author}{\bibinfo{person}{Joseph~David Daron}, \bibinfo{person}{Kate
  Sutherland}, \bibinfo{person}{Christopher Jack}, {and}
  \bibinfo{person}{Bruce~C. Hewitson}.} \bibinfo{year}{2015}\natexlab{}.
\newblock \showarticletitle{The role of regional climate projections in
  managing complex socio-ecological systems}.
\newblock \bibinfo{journal}{\emph{Regional Environmental Change}}
  \bibinfo{volume}{15}, \bibinfo{number}{1} (\bibinfo{date}{01 Jan}
  \bibinfo{year}{2015}), \bibinfo{pages}{1--12}.
\newblock
\urldef\tempurl%
\url{https://doi.org/10.1007/s10113-014-0631-y}
\showDOI{\tempurl}


\bibitem[\protect\citeauthoryear{Davies, Edmondson, Heinemeyer, Leake, and
  Gaston}{Davies et~al\mbox{.}}{2011}]%
        {davies2011mapping}
\bibfield{author}{\bibinfo{person}{Zoe~G Davies}, \bibinfo{person}{Jill~L
  Edmondson}, \bibinfo{person}{Andreas Heinemeyer}, \bibinfo{person}{Jonathan~R
  Leake}, {and} \bibinfo{person}{Kevin~J Gaston}.}
  \bibinfo{year}{2011}\natexlab{}.
\newblock \showarticletitle{Mapping an urban ecosystem service: quantifying
  above-ground carbon storage at a city-wide scale}.
\newblock \bibinfo{journal}{\emph{Journal of applied ecology}}
  \bibinfo{volume}{48}, \bibinfo{number}{5} (\bibinfo{year}{2011}),
  \bibinfo{pages}{1125--1134}.
\newblock


\bibitem[\protect\citeauthoryear{Drusch, Del~Bello, Carlier, Colin, Fernandez,
  Gascon, Hoersch, Isola, Laberinti, Martimort, et~al\mbox{.}}{Drusch
  et~al\mbox{.}}{2012}]%
        {drusch2012sentinel}
\bibfield{author}{\bibinfo{person}{Matthias Drusch}, \bibinfo{person}{Umberto
  Del~Bello}, \bibinfo{person}{S{\'e}bastien Carlier}, \bibinfo{person}{Olivier
  Colin}, \bibinfo{person}{Veronica Fernandez}, \bibinfo{person}{Ferran
  Gascon}, \bibinfo{person}{Bianca Hoersch}, \bibinfo{person}{Claudia Isola},
  \bibinfo{person}{Paolo Laberinti}, \bibinfo{person}{Philippe Martimort},
  {et~al\mbox{.}}} \bibinfo{year}{2012}\natexlab{}.
\newblock \showarticletitle{Sentinel-2: ESA's optical high-resolution mission
  for GMES operational services}.
\newblock \bibinfo{journal}{\emph{Remote sensing of Environment}}
  \bibinfo{volume}{120} (\bibinfo{year}{2012}), \bibinfo{pages}{25--36}.
\newblock


\bibitem[\protect\citeauthoryear{Edenhofer}{Edenhofer}{2015}]%
        {edenhofer2015climate}
\bibfield{author}{\bibinfo{person}{Ottmar Edenhofer}.}
  \bibinfo{year}{2015}\natexlab{}.
\newblock \bibinfo{booktitle}{\emph{Climate change 2014: mitigation of climate
  change}}. Vol.~\bibinfo{volume}{3}.
\newblock \bibinfo{publisher}{Cambridge University Press}.
\newblock


\bibitem[\protect\citeauthoryear{Eggleston, Buendia, Miwa, Ngara, and
  Tanabe}{Eggleston et~al\mbox{.}}{2006}]%
        {eggleston2006}
\bibfield{author}{\bibinfo{person}{Simon Eggleston}, \bibinfo{person}{Leandro
  Buendia}, \bibinfo{person}{Kyoko Miwa}, \bibinfo{person}{Todd Ngara}, {and}
  \bibinfo{person}{Kiyoto Tanabe}.} \bibinfo{year}{2006}\natexlab{}.
\newblock \bibinfo{booktitle}{\emph{2006 IPCC guidelines for national
  greenhouse gas inventories}}. Vol.~\bibinfo{volume}{5}.
\newblock \bibinfo{publisher}{Institute for Global Environmental Strategies
  Hayama, Japan}.
\newblock


\bibitem[\protect\citeauthoryear{EPA}{EPA}{2019}]%
        {epa_passenger_emissions}
\bibfield{author}{\bibinfo{person}{United States Environmental
  Protection~Agency EPA}.} \bibinfo{year}{2019}\natexlab{}.
\newblock \bibinfo{title}{Greenhouse Gas Emissions from a Typical Passenger
  Vehicle}.
\newblock
\newblock
\urldef\tempurl%
\url{https://www.epa.gov/greenvehicles/greenhouse-gas-emissions-typical-passenger-vehicle}
\showURL{%
\tempurl}


\bibitem[\protect\citeauthoryear{Figueres, Schellnhuber, Whiteman,
  Rockstr{\"o}m, Hobley, and Rahmstorf}{Figueres et~al\mbox{.}}{2017}]%
        {figueres2017three}
\bibfield{author}{\bibinfo{person}{Christiana Figueres},
  \bibinfo{person}{Hans~Joachim Schellnhuber}, \bibinfo{person}{Gail Whiteman},
  \bibinfo{person}{Johan Rockstr{\"o}m}, \bibinfo{person}{Anthony Hobley},
  {and} \bibinfo{person}{Stefan Rahmstorf}.} \bibinfo{year}{2017}\natexlab{}.
\newblock \showarticletitle{Three years to safeguard our climate}.
\newblock \bibinfo{journal}{\emph{Nature News}} \bibinfo{volume}{546},
  \bibinfo{number}{7660} (\bibinfo{year}{2017}), \bibinfo{pages}{593}.
\newblock


\bibitem[\protect\citeauthoryear{{GDAL/OGR contributors}}{{GDAL/OGR
  contributors}}{2019}]%
        {gdalcite}
\bibfield{author}{\bibinfo{person}{{GDAL/OGR contributors}}.}
  \bibinfo{year}{2019}\natexlab{}.
\newblock \bibinfo{booktitle}{\emph{{GDAL/OGR} Geospatial Data Abstraction
  software Library}}.
\newblock Open Source Geospatial Foundation.
\newblock
\urldef\tempurl%
\url{https://gdal.org}
\showURL{%
\tempurl}


\bibitem[\protect\citeauthoryear{Gil, Deelman, Demir, Duffy, Pierce, Marru, and
  Wiener}{Gil et~al\mbox{.}}{2012a}]%
        {gil-etal-agu12}
\bibfield{author}{\bibinfo{person}{Yolanda Gil}, \bibinfo{person}{Ewa Deelman},
  \bibinfo{person}{Ibrahim Demir}, \bibinfo{person}{Chris Duffy},
  \bibinfo{person}{Marlon Pierce}, \bibinfo{person}{Suresh Marru}, {and}
  \bibinfo{person}{Gerry Wiener}.} \bibinfo{year}{2012}\natexlab{a}.
\newblock \bibinfo{title}{Designing a Roadmap for Workflow Cyberinfrastructure
  in the Geosciences: From Big Data to the Long Tail}.
\newblock
\newblock
\newblock
\shownote{American Geophysical Union Fall Meeting, San Francisco, CA.}


\bibitem[\protect\citeauthoryear{Gil, Ratnakar, Deelman, and Mason}{Gil
  et~al\mbox{.}}{2012b}]%
        {gil-etal-ismb12}
\bibfield{author}{\bibinfo{person}{Yolanda Gil}, \bibinfo{person}{Varun
  Ratnakar}, \bibinfo{person}{Ewa Deelman}, {and} \bibinfo{person}{Christopher
  Mason}.} \bibinfo{year}{2012}\natexlab{b}.
\newblock \bibinfo{title}{Using Semantic Workflows for Genome-Scale Analysis}.
\newblock
\newblock
\newblock
\shownote{International Conference on Intelligent Systems for Molecular Biology
  (ISMB), Long Beach, CA.}


\bibitem[\protect\citeauthoryear{Gil, Ratnakar, Kim, Gonzalez-Calero, Groth,
  Moody, and Deelman}{Gil et~al\mbox{.}}{2011}]%
        {gil-etal-ieee-is-11}
\bibfield{author}{\bibinfo{person}{Yolanda Gil}, \bibinfo{person}{Varun
  Ratnakar}, \bibinfo{person}{Jihie Kim}, \bibinfo{person}{Pedro~Antonio
  Gonzalez-Calero}, \bibinfo{person}{Paul Groth}, \bibinfo{person}{Joshua
  Moody}, {and} \bibinfo{person}{Ewa Deelman}.}
  \bibinfo{year}{2011}\natexlab{}.
\newblock \showarticletitle{Wings: Intelligent Workflow-Based Design of
  Computational Experiments}.
\newblock \bibinfo{journal}{\emph{IEEE Intelligent Systems}}
  \bibinfo{volume}{26}, \bibinfo{number}{1} (\bibinfo{year}{2011}).
\newblock
\urldef\tempurl%
\url{http://www.isi.edu/~gil/papers/gil-etal-ieee-is-11.pdf}
\showURL{%
\tempurl}


\bibitem[\protect\citeauthoryear{Gillingham, Nordhaus, Anthoff, Blanford,
  Bosetti, Christensen, McJeon, and Reilly}{Gillingham et~al\mbox{.}}{2018}]%
        {gillingham2018modeling}
\bibfield{author}{\bibinfo{person}{Kenneth Gillingham},
  \bibinfo{person}{William Nordhaus}, \bibinfo{person}{David Anthoff},
  \bibinfo{person}{Geoffrey Blanford}, \bibinfo{person}{Valentina Bosetti},
  \bibinfo{person}{Peter Christensen}, \bibinfo{person}{Haewon McJeon}, {and}
  \bibinfo{person}{John Reilly}.} \bibinfo{year}{2018}\natexlab{}.
\newblock \showarticletitle{Modeling uncertainty in integrated assessment of
  climate change: A multimodel comparison}.
\newblock \bibinfo{journal}{\emph{Journal of the Association of Environmental
  and Resource Economists}} \bibinfo{volume}{5}, \bibinfo{number}{4}
  (\bibinfo{year}{2018}), \bibinfo{pages}{791--826}.
\newblock


\bibitem[\protect\citeauthoryear{Gough}{Gough}{2016}]%
        {gough2016carbon}
\bibfield{author}{\bibinfo{person}{Clair Gough}.}
  \bibinfo{year}{2016}\natexlab{}.
\newblock \bibinfo{booktitle}{\emph{Carbon capture and its storage: an
  integrated assessment}}.
\newblock \bibinfo{publisher}{Routledge}.
\newblock


\bibitem[\protect\citeauthoryear{Grizonnet, Michel, Poughon, Inglada, Savinaud,
  and Cresson}{Grizonnet et~al\mbox{.}}{2017}]%
        {grizonnet2017orfeo}
\bibfield{author}{\bibinfo{person}{Manuel Grizonnet}, \bibinfo{person}{Julien
  Michel}, \bibinfo{person}{Victor Poughon}, \bibinfo{person}{Jordi Inglada},
  \bibinfo{person}{Micka{\"e}l Savinaud}, {and} \bibinfo{person}{R{\'e}mi
  Cresson}.} \bibinfo{year}{2017}\natexlab{}.
\newblock \showarticletitle{Orfeo ToolBox: Open source processing of remote
  sensing images}.
\newblock \bibinfo{journal}{\emph{Open Geospatial Data, Software and
  Standards}} \bibinfo{volume}{2}, \bibinfo{number}{1} (\bibinfo{year}{2017}),
  \bibinfo{pages}{15}.
\newblock


\bibitem[\protect\citeauthoryear{Haklay and Weber}{Haklay and Weber}{2008}]%
        {haklay2008openstreetmap}
\bibfield{author}{\bibinfo{person}{Mordechai Haklay} {and}
  \bibinfo{person}{Patrick Weber}.} \bibinfo{year}{2008}\natexlab{}.
\newblock \showarticletitle{Openstreetmap: User-generated street maps}.
\newblock \bibinfo{journal}{\emph{IEEE Pervasive Computing}}
  \bibinfo{volume}{7}, \bibinfo{number}{4} (\bibinfo{year}{2008}),
  \bibinfo{pages}{12--18}.
\newblock


\bibitem[\protect\citeauthoryear{Incropera}{Incropera}{2016}]%
        {incropera2016climate}
\bibfield{author}{\bibinfo{person}{Frank~P Incropera}.}
  \bibinfo{year}{2016}\natexlab{}.
\newblock \bibinfo{booktitle}{\emph{Climate change: a wicked problem:
  complexity and uncertainty at the intersection of science, economics,
  politics, and human behavior}}.
\newblock \bibinfo{publisher}{Cambridge University Press}.
\newblock


\bibitem[\protect\citeauthoryear{Kagawa and Le~Sourda}{Kagawa and
  Le~Sourda}{2018}]%
        {kagawa2018mapping}
\bibfield{author}{\bibinfo{person}{Ayako Kagawa} {and}
  \bibinfo{person}{Guillaume Le~Sourda}.} \bibinfo{year}{2018}\natexlab{}.
\newblock \showarticletitle{Mapping the world: cartographic and geographic
  visualization by the United Nations Geospatial Information Section (formerly
  Cartographic Section)}. In \bibinfo{booktitle}{\emph{Proceedings of the
  ICA}}, Vol.~\bibinfo{volume}{1}. \bibinfo{pages}{58}.
\newblock


\bibitem[\protect\citeauthoryear{Kundzewicz, Krysanova, Benestad, Hov,
  Piniewski, and Otto}{Kundzewicz et~al\mbox{.}}{2018}]%
        {kundzewicz2018uncertainty}
\bibfield{author}{\bibinfo{person}{ZW Kundzewicz}, \bibinfo{person}{V
  Krysanova}, \bibinfo{person}{RE Benestad}, \bibinfo{person}{{\O} Hov},
  \bibinfo{person}{M Piniewski}, {and} \bibinfo{person}{IM Otto}.}
  \bibinfo{year}{2018}\natexlab{}.
\newblock \showarticletitle{Uncertainty in climate change impacts on water
  resources}.
\newblock \bibinfo{journal}{\emph{Environmental Science \& Policy}}
  \bibinfo{volume}{79} (\bibinfo{year}{2018}), \bibinfo{pages}{1--8}.
\newblock


\bibitem[\protect\citeauthoryear{Livesley, McPherson, and Calfapietra}{Livesley
  et~al\mbox{.}}{2016}]%
        {livesley2016urban}
\bibfield{author}{\bibinfo{person}{SJ Livesley}, \bibinfo{person}{EG
  McPherson}, {and} \bibinfo{person}{C Calfapietra}.}
  \bibinfo{year}{2016}\natexlab{}.
\newblock \showarticletitle{The urban forest and ecosystem services: impacts on
  urban water, heat, and pollution cycles at the tree, street, and city scale}.
\newblock \bibinfo{journal}{\emph{Journal of environmental quality}}
  \bibinfo{volume}{45}, \bibinfo{number}{1} (\bibinfo{year}{2016}),
  \bibinfo{pages}{119--124}.
\newblock


\bibitem[\protect\citeauthoryear{Macharia, Ouma, Gogo, Snow, and Noor}{Macharia
  et~al\mbox{.}}{2017}]%
        {macharia2017spatial}
\bibfield{author}{\bibinfo{person}{Peter~M Macharia}, \bibinfo{person}{Paul~O
  Ouma}, \bibinfo{person}{Ezekiel~G Gogo}, \bibinfo{person}{Robert~W Snow},
  {and} \bibinfo{person}{Abdisalan~M Noor}.} \bibinfo{year}{2017}\natexlab{}.
\newblock \showarticletitle{Spatial accessibility to basic public health
  services in South Sudan}.
\newblock \bibinfo{journal}{\emph{Geospatial health}} \bibinfo{volume}{12},
  \bibinfo{number}{1} (\bibinfo{year}{2017}), \bibinfo{pages}{510}.
\newblock


\bibitem[\protect\citeauthoryear{McGovern and Pasher}{McGovern and
  Pasher}{2016}]%
        {mcgovern2016canadian}
\bibfield{author}{\bibinfo{person}{Mark McGovern} {and} \bibinfo{person}{Jon
  Pasher}.} \bibinfo{year}{2016}\natexlab{}.
\newblock \showarticletitle{Canadian urban tree canopy cover and carbon
  sequestration status and change 1990--2012}.
\newblock \bibinfo{journal}{\emph{Urban forestry \& urban greening}}
  \bibinfo{volume}{20} (\bibinfo{year}{2016}), \bibinfo{pages}{227--232}.
\newblock


\bibitem[\protect\citeauthoryear{Meadow, Ferguson, Guido, Horangic, Owen, and
  Wall}{Meadow et~al\mbox{.}}{2015}]%
        {meadow2015moving}
\bibfield{author}{\bibinfo{person}{Alison~M Meadow}, \bibinfo{person}{Daniel~B
  Ferguson}, \bibinfo{person}{Zack Guido}, \bibinfo{person}{Alexandra
  Horangic}, \bibinfo{person}{Gigi Owen}, {and} \bibinfo{person}{Tamara Wall}.}
  \bibinfo{year}{2015}\natexlab{}.
\newblock \showarticletitle{Moving toward the deliberate coproduction of
  climate science knowledge}.
\newblock \bibinfo{journal}{\emph{Weather, Climate, and Society}}
  \bibinfo{volume}{7}, \bibinfo{number}{2} (\bibinfo{year}{2015}),
  \bibinfo{pages}{179--191}.
\newblock


\bibitem[\protect\citeauthoryear{Nassif, Stewart, Mutepfe, and Christou}{Nassif
  et~al\mbox{.}}{2016}]%
        {nassif2016infrastructure}
\bibfield{author}{\bibinfo{person}{Ayman Nassif}, \bibinfo{person}{Ann
  Stewart}, \bibinfo{person}{Millie Mutepfe}, {and} \bibinfo{person}{Petros
  Christou}.} \bibinfo{year}{2016}\natexlab{}.
\newblock \showarticletitle{Infrastructure needs in sub Saharan Africa with
  particular reference to South Sudan}. In \bibinfo{booktitle}{\emph{Third
  Australasia and South-East Asia Structural Engineering and Construction
  Conference: ASEA-SEC-3}}. ISEC Press.
\newblock


\bibitem[\protect\citeauthoryear{Nowak, Greenfield, Hoehn, and Lapoint}{Nowak
  et~al\mbox{.}}{2013}]%
        {nowak2013carbon}
\bibfield{author}{\bibinfo{person}{David~J Nowak}, \bibinfo{person}{Eric~J
  Greenfield}, \bibinfo{person}{Robert~E Hoehn}, {and}
  \bibinfo{person}{Elizabeth Lapoint}.} \bibinfo{year}{2013}\natexlab{}.
\newblock \showarticletitle{Carbon storage and sequestration by trees in urban
  and community areas of the United States}.
\newblock \bibinfo{journal}{\emph{Environmental pollution}}
  \bibinfo{volume}{178} (\bibinfo{year}{2013}), \bibinfo{pages}{229--236}.
\newblock


\bibitem[\protect\citeauthoryear{Pasher, McGovern, Khoury, and Duffe}{Pasher
  et~al\mbox{.}}{2014}]%
        {pasher2014assessing}
\bibfield{author}{\bibinfo{person}{Jon Pasher}, \bibinfo{person}{Mark
  McGovern}, \bibinfo{person}{Michael Khoury}, {and} \bibinfo{person}{Jason
  Duffe}.} \bibinfo{year}{2014}\natexlab{}.
\newblock \showarticletitle{Assessing carbon storage and sequestration by
  Canada's urban forests using high resolution earth observation data}.
\newblock \bibinfo{journal}{\emph{Urban forestry \& urban greening}}
  \bibinfo{volume}{13}, \bibinfo{number}{3} (\bibinfo{year}{2014}),
  \bibinfo{pages}{484--494}.
\newblock


\bibitem[\protect\citeauthoryear{Penman, Gytarsky, Hiraishi, Krug, Kruger,
  Pipatti, Buendia, Miwa, Ngara, Tanabe, et~al\mbox{.}}{Penman
  et~al\mbox{.}}{2003}]%
        {penman2003good}
\bibfield{author}{\bibinfo{person}{Jim Penman}, \bibinfo{person}{Michael
  Gytarsky}, \bibinfo{person}{Taka Hiraishi}, \bibinfo{person}{Thelma Krug},
  \bibinfo{person}{Dina Kruger}, \bibinfo{person}{Riitta Pipatti},
  \bibinfo{person}{Leandro Buendia}, \bibinfo{person}{Kyoko Miwa},
  \bibinfo{person}{Todd Ngara}, \bibinfo{person}{Kiyoto Tanabe},
  {et~al\mbox{.}}} \bibinfo{year}{2003}\natexlab{}.
\newblock \showarticletitle{Good practice guidance for land use, land-use
  change and forestry.}
\newblock \bibinfo{journal}{\emph{Good practice guidance for land use, land-use
  change and forestry.}} (\bibinfo{year}{2003}).
\newblock


\bibitem[\protect\citeauthoryear{Raciti, Hutyra, and Newell}{Raciti
  et~al\mbox{.}}{2014}]%
        {raciti2014mapping}
\bibfield{author}{\bibinfo{person}{Steve~M Raciti}, \bibinfo{person}{Lucy~R
  Hutyra}, {and} \bibinfo{person}{Jared~D Newell}.}
  \bibinfo{year}{2014}\natexlab{}.
\newblock \showarticletitle{Mapping carbon storage in urban trees with
  multi-source remote sensing data: Relationships between biomass, land use,
  and demographics in Boston neighborhoods}.
\newblock \bibinfo{journal}{\emph{Science of the Total Environment}}
  \bibinfo{volume}{500} (\bibinfo{year}{2014}), \bibinfo{pages}{72--83}.
\newblock


\bibitem[\protect\citeauthoryear{Schreyer, Tigges, Lakes, and
  Churkina}{Schreyer et~al\mbox{.}}{2014}]%
        {schreyer2014using}
\bibfield{author}{\bibinfo{person}{Johannes Schreyer}, \bibinfo{person}{Jan
  Tigges}, \bibinfo{person}{Tobia Lakes}, {and} \bibinfo{person}{Galina
  Churkina}.} \bibinfo{year}{2014}\natexlab{}.
\newblock \showarticletitle{Using airborne LiDAR and QuickBird data for
  modelling urban tree carbon storage and its distribution—A case study of
  Berlin}.
\newblock \bibinfo{journal}{\emph{Remote Sensing}} \bibinfo{volume}{6},
  \bibinfo{number}{11} (\bibinfo{year}{2014}), \bibinfo{pages}{10636--10655}.
\newblock


\bibitem[\protect\citeauthoryear{Schulson}{Schulson}{2018}]%
        {wired}
\bibfield{author}{\bibinfo{person}{Michael Schulson}.}
  \bibinfo{year}{2018}\natexlab{}.
\newblock \showarticletitle{Science's Reproducibility Crisis Is Being Used as
  Political Ammunition}.
\newblock \bibinfo{journal}{\emph{Wired}} (\bibinfo{year}{2018}).
\newblock
\urldef\tempurl%
\url{https://www.wired.com/story/sciences-reproducibility-crisis-is-being-used-as-political-ammunition/}
\showURL{%
\tempurl}


\bibitem[\protect\citeauthoryear{S{\o}rb{\o} and Ahmed}{S{\o}rb{\o} and
  Ahmed}{2013}]%
        {sorbo2013sudan}
\bibfield{author}{\bibinfo{person}{Gunnar~M S{\o}rb{\o}} {and}
  \bibinfo{person}{Abdel Ghaffar~M Ahmed}.} \bibinfo{year}{2013}\natexlab{}.
\newblock \bibinfo{booktitle}{\emph{Sudan divided: continuing conflict in a
  contested state}}.
\newblock \bibinfo{publisher}{Springer}.
\newblock


\bibitem[\protect\citeauthoryear{Taylor, Deelman, Gannon, and Shields}{Taylor
  et~al\mbox{.}}{2014}]%
        {Taylor:2014}
\bibfield{author}{\bibinfo{person}{Ian~J. Taylor}, \bibinfo{person}{Ewa
  Deelman}, \bibinfo{person}{Dennis~B. Gannon}, {and} \bibinfo{person}{Matthew
  Shields}.} \bibinfo{year}{2014}\natexlab{}.
\newblock \bibinfo{booktitle}{\emph{Workflows for e-Science: Scientific
  Workflows for Grids}}.
\newblock \bibinfo{publisher}{Springer Publishing Company, Incorporated}.
\newblock
\showISBNx{1849966192, 9781849966191}


\bibitem[\protect\citeauthoryear{Tigges, Churkina, and Lakes}{Tigges
  et~al\mbox{.}}{2017}]%
        {tigges2017modeling}
\bibfield{author}{\bibinfo{person}{Jan Tigges}, \bibinfo{person}{Galina
  Churkina}, {and} \bibinfo{person}{Tobia Lakes}.}
  \bibinfo{year}{2017}\natexlab{}.
\newblock \showarticletitle{Modeling above-ground carbon storage: a remote
  sensing approach to derive individual tree species information in urban
  settings}.
\newblock \bibinfo{journal}{\emph{Urban ecosystems}} \bibinfo{volume}{20},
  \bibinfo{number}{1} (\bibinfo{year}{2017}), \bibinfo{pages}{97--111}.
\newblock


\bibitem[\protect\citeauthoryear{Wolstencroft, Haines, Fellows, Williams,
  Withers, Owen, Soiland-Reyes, Dunlop, Nenadic, Fisher,
  et~al\mbox{.}}{Wolstencroft et~al\mbox{.}}{2013}]%
        {wolstencroft2013taverna}
\bibfield{author}{\bibinfo{person}{Katherine Wolstencroft},
  \bibinfo{person}{Robert Haines}, \bibinfo{person}{Donal Fellows},
  \bibinfo{person}{Alan Williams}, \bibinfo{person}{David Withers},
  \bibinfo{person}{Stuart Owen}, \bibinfo{person}{Stian Soiland-Reyes},
  \bibinfo{person}{Ian Dunlop}, \bibinfo{person}{Aleksandra Nenadic},
  \bibinfo{person}{Paul Fisher}, {et~al\mbox{.}}}
  \bibinfo{year}{2013}\natexlab{}.
\newblock \showarticletitle{The Taverna workflow suite: designing and executing
  workflows of Web Services on the desktop, web or in the cloud}.
\newblock \bibinfo{journal}{\emph{Nucleic acids research}}
  \bibinfo{volume}{41}, \bibinfo{number}{W1} (\bibinfo{year}{2013}),
  \bibinfo{pages}{W557--W561}.
\newblock


\end{thebibliography}

\end{document}